\documentclass[letterpaper, 10 pt, conference]{ieeeconf}  

\IEEEoverridecommandlockouts                              

\usepackage{graphicx} 
\usepackage{booktabs}
\usepackage{multirow}
\usepackage{amssymb}
\usepackage{amsmath} 
\usepackage[nocompress]{cite} 

\makeatletter
\let\NAT@parse\undefined
\makeatother
\usepackage[colorlinks=true, linkcolor=black, citecolor=blue, urlcolor=blue, breaklinks=true]{hyperref}

\title{\LARGE \bf
RADAR: Closed-Loop Robotic Data Generation via Semantic Planning and Autonomous Causal Environment Reset
}

\author{
Yongzhong Wang$^{1,*}$, Keyu Zhu$^{1,*}$, Yong Zhong$^{1,*}$, \
Liqiong Wang$^{1}$, Jinyu Yang$^{2,\dagger}$, Feng Zheng$^{1,3}$ \
\vspace{0.15cm}
\thanks{$^{*}$Equal contribution.}
\thanks{$^{\dagger}$Corresponding author.}
\thanks{$^{1}$Department of Computer Science and Engineering, Southern University of Science and Technology.}%
\thanks{$^{2}$Harbin Institute of Technology, Shenzhen.}%
\thanks{$^{3}$Spatialtemporal AI.}%
}

\begin{document}

\maketitle
\thispagestyle{empty}
\pagestyle{empty}


\begin{figure*}[t!]
    \centering
    \includegraphics[width=0.9\linewidth]{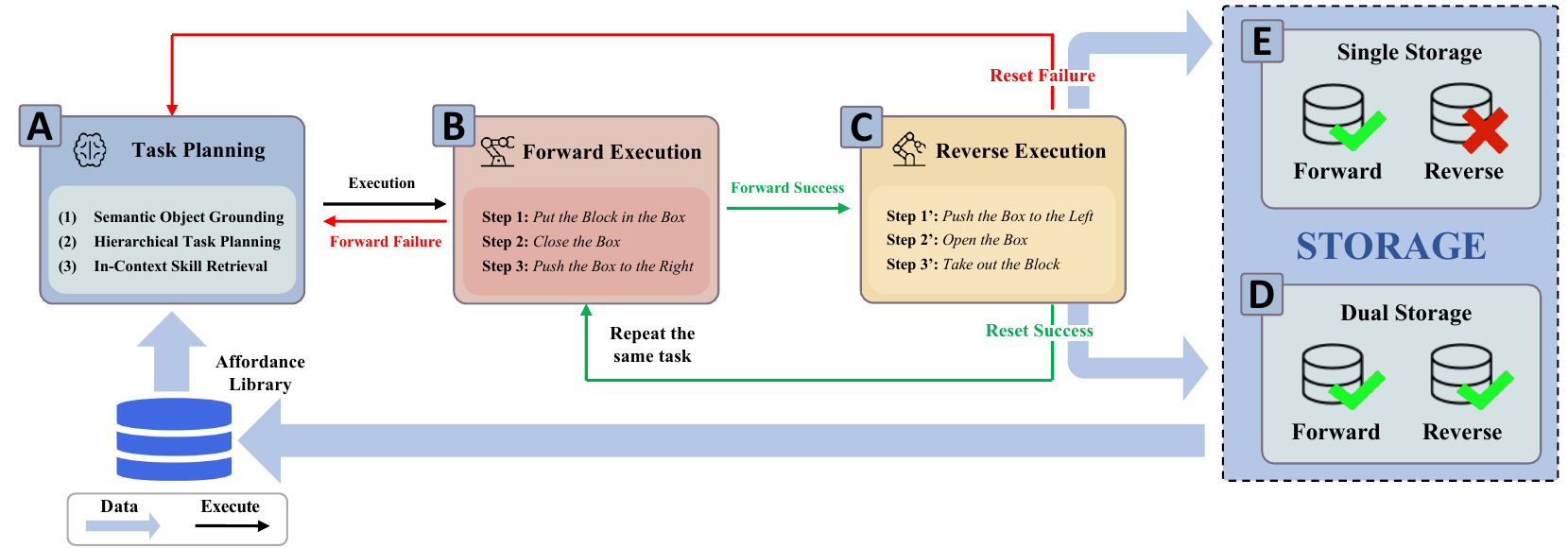} 
    \caption{Overview of the RADAR pipeline and the state transition diagram of its decoupled Finite State Machine (FSM). To ensure logical clarity, the architecture strictly separates physical execution loops (\textbf{States A, B, C}) from concurrent data routing actions (\textbf{States D, E}). A fully successful execution forms a continuous loop (\textbf{B $\to$ C $\to$ B}), concurrently triggering \textbf{Dual Storage (D)} to repeatedly harvest trajectory variations without re-planning. In contrast, an asymmetric recovery loop (\textbf{B $\to$ C $\to$ A}) bypasses reset failures by selectively saving the valid forward trajectory via \textbf{Single Storage (E)} and initiating a novel planning cycle on the altered workspace. This architecture guarantees a truly self-sustaining, human-out-of-the-loop engine.}
    \label{fig:whole_logic}
\end{figure*}

\begin{abstract}
The acquisition of large-scale physical interaction data—a critical prerequisite for modern robot learning—is severely bottlenecked by the prohibitive cost and scalability limits of human-in-the-loop collection paradigms. To break this barrier, we introduce Robust Autonomous Data Acquisition for Robotics (RADAR), a fully autonomous, closed-loop data generation engine that completely removes human intervention from the collection cycle. RADAR elegantly divides the cognitive load into a four-module pipeline. Anchored by merely 2-5 3D human demonstrations as geometric priors, a Vision-Language Model (VLM) first orchestrates scene-relevant task generation via precise semantic object grounding and skill retrieval. Next, a Graph Neural Network (GNN) policy translates these subtasks into robust physical actions via in-context imitation learning. Following execution, the VLM performs automated success evaluation using a structured Visual Question Answering (VQA) pipeline. Finally, to shatter the bottleneck of manual resets, a Finite State Machine (FSM) orchestrates an autonomous environment reset and asymmetric data routing mechanism. Driven by simultaneous forward-reverse planning with a strict Last-In, First-Out (LIFO) causal sequence, the system seamlessly restores unstructured workspaces and robustly recovers from execution failures. This continuous brain-cerebellum synergy transforms data collection into a self-sustaining process. Extensive evaluations highlight RADAR's exceptional versatility. In simulation, our framework achieves up to 90\% success rates on complex, long-horizon tasks, effortlessly solving challenges where traditional baselines plummet to near-zero performance. During real-world deployments, the system reliably executes diverse, contact-rich atomic skills—such as deformable object manipulation—via few-shot adaptation without domain-specific fine-tuning, providing a highly scalable paradigm to democratize robotic data acquisition. More demonstration are available on our
\href{https://radar-iros.netlify.app/}{project page}.

\end{abstract}

\section{Introduction}

Recent end-to-end embodied intelligence models have demonstrated remarkable generalization capabilities, driving significant progress in robotic manipulation~\cite{black2024pi0,physicalintelligence2025pi05,liu2024rdt1b,yang2026abot}. However, the scaling of such models is fundamentally bottlenecked by the acquisition of large-scale, high-fidelity physical interaction data. Existing solutions to this data bottleneck face a frustrating dichotomy. Simulation-based methods~\cite{dalal2023imitating,wang2023robogen,mandlekar2023mimicgen,jiang2024dexmimicen} offer tremendous scalability but struggle with persistent sim-to-real gaps and limited behavioral diversity. Conversely, teleoperation-based methods~\cite{brohan2022rt1,ebert2022bridge,brohan2023say,lynch2023interactive} provide high-quality demonstrations but remain prohibitively expensive and fundamentally unscalable due to the slow and serial nature of human control.


Recent advances have addressed several individual components required for
autonomous robot data generation, yet these components remain largely
disconnected. At the planning level, VLM-based manipulation methods represent and organize intended behaviors through 2D visual prompts, generated image subgoals,
geometric constraints, or agentic task graphs~\cite{fang2024moka,huang2024rekep,zhou2024autonomous,black2023susie,xu2026reaction},
but such representations provide limited geometric constraints for precise
physical interaction. At the execution level, In-Context Imitation Learning
and generative visuomotor policies enable accurate control from demonstrations
~\cite{vosylius2024instantpolicy,chi2023diffusion,goo2023vid2robot},
but typically assume that task specifications, execution contexts, and
demonstrations are supplied externally. Meanwhile, autonomous learning
frameworks~\cite{zhou2024autonomous,bousmalis2023robocat} automate parts of task
proposal, evaluation, or policy improvement, but do not provide a general
mechanism for restoring the physical workspace after interaction. Consequently,
existing approaches do not yet form a self-sustaining collection loop that
jointly supports task generation, precise execution, outcome verification,
and autonomous environment reset.


To address these limitations, we introduce \textbf{R}obust \textbf{A}utonomous \textbf{D}ata \textbf{A}cquisition for \textbf{R}obotics (\textbf{RADAR}), a fully automated, closed-loop pipeline, as illustrated in Fig.~\ref{fig:whole_logic}. Rather than requiring Vision-Language Models (VLMs) to generate intermediate images or directly infer fragile 3D coordinates, or relying on disjointed heuristic scripts for environment reset, RADAR establishes a \textbf{brain--cerebellum synergy} by separating high-level cognitive reasoning from low-level physical execution. Specifically, the VLM serves as the cognitive ``brain,'' responsible for semantic reasoning, task generation, and success evaluation, while the GNN-based policy acts as the ``cerebellum,'' executing sub-millimeter, high-frequency control based on 3D geometric priors. Building on recent advances in In-Context Learning (ICL)~\cite{jang2022bcz,vosylius2024instantpolicy}, RADAR uses a small number of human demonstrations as reusable 3D physical priors, enabling the system to generalize these demonstrations into large-scale physical execution. In this way, the framework reduces the sim-to-real gap associated with simulation, the teleoperation burden of manual data collection, and the geometric hallucination risks of direct VLM control. \textbf{As a result, RADAR enables the continuous acquisition of high-fidelity robotic data with minimal human intervention.} The complete data-generation process is organized into four modules:

(1) \textbf{Scene-Relevant Task Generation}: We employ a Vision-Language Model (VLM) to autonomously construct scene-relevant tasks and extract object segmentation masks based on the current observation. Complex, long-horizon tasks are decomposed into a sequence of atomic subtasks, each matched with a relevant demonstration from a small affordance library as a behavioral prior.

(2) \textbf{Task Execution via In-Context Imitation Learning}: The robot performs the assigned subtasks using a Graph Neural Network (GNN)-based in-context imitation learning framework, which maps the selected demonstrations and current observations to executable continuous trajectories.

(3) \textbf{Automated Success Evaluation}: The VLM acts as an embodied evaluator to determine the outcome of the execution through a structured Visual Question Answering (VQA) formulation, filtering out failed trajectories.

(4) \textbf{Autonomous Environment Reset}: Upon completion of a task, a Finite State Machine (FSM) governs the system to actively compute and execute a causal inverse sequence of the forward actions. This strictly adheres to a Last-In, First-Out (LIFO) logical constraint to restore the environment, facilitating truly continuous and robust data generation \textbf{without human assistance}.


While drastically reducing reliance on human labor, RADAR maintains strong
performance across both simulated and real-world deployments. In simulation,
we evaluate the framework's ability to plan and execute complex, long-horizon
manipulation tasks, where it achieves high success rates and robust autonomous
resetting. We further validate its practical applicability through deployment
on a physical robotic system. Using only one or a few visual demonstrations,
RADAR successfully performs a range of challenging atomic skills, including
deformable-object manipulation, such as towel folding, and high-precision
alignment, such as paper-roll insertion, without requiring domain-specific
fine-tuning. Taken together, these results demonstrate the robustness and
versatility of our closed-loop pipeline across different domains, establishing
RADAR as a scalable and domain-agnostic data-generation engine for physical
robot learning.

Our main contributions are highlighted as follows:
\begin{itemize}
    \item We introduce \textbf{RADAR}, a \textbf{fully automated, closed-loop} pipeline for real-world robot manipulation data collection. Our system requires only 2-5 manually collected atomic demonstrations and scales them into diverse, task-relevant datasets with strictly minimal human intervention.
    
    \item We propose a \textbf{scene-relevant task generation} framework, which effectively translates complex, long-horizon tasks in cluttered environments into sequentially executable atomic skills.
    
    \item We design a novel \textbf{autonomous environment reset mechanism} to achieve autonomous environment resetting. This empowers the robot with causal self-correction and scene-restoration capabilities, unlocking continuous, human-out-of-the-loop data streaming.

    \item \textbf{Extensive Empirical Validation:} We validate RADAR's capabilities across simulated and physical domains. Our pipeline achieves up to 90\% success rates on complex, long-horizon simulated tasks, demonstrating highly robust forward planning and execution. Furthermore, real-world deployments establish the system as a powerful proof-of-concept for human-out-of-the-loop data generation, reliably executing diverse, contact-rich physical skills (\textit{e.g.}, deformable object manipulation) via few-shot in-context learning.
\end{itemize}

\section{RELATED WORK}

\subsection{Autonomous Data Collection and Evaluation}
\label{subsec:autonomous_data_collection}

High-quality robot demonstration data is the cornerstone of robust imitation learning \cite{brohan2022rt1, padalkar2023openx}. To overcome manual collection bottlenecks, recent works explore autonomous policy improvement \cite{bousmalis2023robocat}. For instance, the SOAR framework \cite{zhou2024autonomous} utilizes pre-trained VLMs for task proposal and success detection, employing an image-editing diffusion model, SuSIE \cite{black2023susie}, to generate visual subgoals for a goal-conditioned policy \cite{andrychowicz2017hindsight}. 

However, SOAR and similar autonomous systems face a fundamental bottleneck: environment resets. Once a robot alters the environment, continuous collection breaks without human intervention. Furthermore, SOAR's single-stage VLM evaluation is highly susceptible to conversational redundancies and visual hallucinations, leading to false-positive success labels \cite{zhou2024autonomous}.

In contrast, our pipeline achieves true closed-loop autonomy via a Simultaneous Forward-Reverse Planning mechanism. The VLM constructs a strict Last-In, First-Out (LIFO) causal sequence to automatically restore the environment. Coupled with an asymmetric failure handling logic, un-restored scenes seamlessly become novel initial states. Moreover, we replace unreliable single-stage evaluation with a robust three-stage Vision-Question-Answering (VQA) pipeline, strictly decoupling VLM visual reasoning from deterministic logic to ensure the absolute fidelity of collected demonstrations.

\subsection{Visual Prompting and Affordance Reasoning}
\label{subsec:visual_prompting}

Integrating VLMs into control architectures requires effective affordance representations \cite{gibson1977theory, manuelli2019kpam}. Inspired by Set-of-Mark prompting \cite{yang2023setofmark}, MOKA \cite{fang2024moka} advanced this via a mark-based visual prompting framework, overlaying 2D candidate keypoints on RGB images to formulate affordance reasoning as a VQA problem \cite{liu2023grounding, fang2024moka}.

While simplifying VLM reasoning, this 2D paradigm exhibits significant geometric vulnerability. Predicting affordances entirely in 2D pixel space forces MOKA to rely on noisy depth heuristics to back-project points into 3D $SE(3)$ space \cite{fang2024moka}. Consequently, executions involving complex contact dynamics (\textit{e.g.}, tight insertions) inevitably fail, as 2D pixels cannot encapsulate precise kinematic constraints.

Our approach firmly rejects this fragile 2D guessing. We introduce a 3D prior-based Affordance Library derived from a small set of real human demonstrations. Instead of forcing the VLM to generate coordinates from scratch, we constrain it to perform semantic object grounding and In-Context Skill Retrieval. By retrieving the most geometrically and semantically consistent 3D demonstration as a contextual prior, we shift the physical precision burden to human-demonstrated trajectories. This preserves VLM semantic generalization while achieving strictly feasible, high-fidelity kinematics.

\subsection{In-Context Learning and Graph Diffusion}
\label{subsec:in_context_learning}

In-Context Imitation Learning (ICIL) has emerged as a promising paradigm to execute VLM-planned subtasks without exhaustive fine-tuning \cite{jang2022bcz, goo2023vid2robot}. Bridging ICIL with multimodal diffusion models \cite{chi2023diffusion}, Instant Policy \cite{vosylius2024instantpolicy} formulates ICIL as a conditional graph generation problem using Graph Neural Networks \cite{wang2018nervenet}. It employs a graph transformer-based reverse diffusion process to iteratively denoise future action nodes \cite{vosylius2024instantpolicy}.

While the underlying graph diffusion architecture is highly effective for low-level action generation, it fundamentally operates as an isolated execution policy. It heavily relies on human-provided task definitions and demonstrations, lacking the cognitive mechanisms required to autonomously propose novel tasks, chain long-horizon behaviors, or verify execution success. Conversely, autonomous frameworks like SOAR attempt to close this loop using image-editing diffusion models (SuSIE) to generate intermediate visual subgoals \cite{black2023susie, zhou2024autonomous}. However, generating intermediate pixels introduces high computational latency and severe physical hallucinations (\textit{e.g.}, floating objects), inevitably causing catastrophic execution failures.

Our work resolves this dichotomy by embedding the graph diffusion model as a subordinate execution engine within a fully autonomous, VLM-driven closed loop. We replace error-prone pixel generation with our robust front-end hierarchical task planning, which autonomously grounds objects and translates high-level semantic goals into precise atomic contexts. Following execution, our back-end VQA evaluation strictly verifies task completion and triggers environment resets. By wrapping the sub-millimeter $\mathbb{SE}(3)$ precision of the ICIL graph architecture with these autonomous cognitive layers, we successfully elevate it from a passive execution policy into a continuous, human-out-of-the-loop data generation pipeline.

\section{METHOD}
\label{sec:method}

As orchestrated by the continuous control flow illustrated in Fig. \ref{fig:whole_logic}, our fully automated data collection framework operates as a self-sustaining closed loop. To ground the system's physical execution capabilities, we first formalize a foundational set of semantic skills—depicted as the vital data prior in our architecture—as an \textbf{Affordance Library}. Building upon this library, our pipeline is elegantly structured into a \textbf{four-module framework} that explicitly mirrors the brain-cerebellum division of labor: (1) Scene-Relevant Task Generation, (2) Task Execution via In-Context Imitation Learning, (3) Automated Success Evaluation, and (4) Autonomous Environment Reset. 

In the following subsections, we first define the prerequisite Affordance Library, and subsequently detail the technical formulation of each of the four pipeline modules.

In the following subsections, we first define the prerequisite Affordance Library, and subsequently detail the technical formulation of each of the four pipeline phases.

\subsection{Preliminaries: Affordance Library}
\label{subsec:affordance_library}

To facilitate downstream in-context imitation learning policy (detailed in Section \ref{subsec:task_execution}), we construct an Affordance Library of manually collected demonstrations. We denote the library as $L$. Following instant policy \cite{vosylius2024instantpolicy}, each demonstration $d \in L$ is a trajectory of local graph capturing the state of execution over a temporal horizon $T$. Formally, a demonstration $d$ is defined as a sequence of states:
\begin{equation}
d = \{(P^t, \mathbf{T}_{WE}^t, g^t, \xi)\}_{t=1}^T
\end{equation}
At each timestamp $t$, our observation at time consists of segmented point cloud $P^t$, the homogeneous transformation matrix from the world frame to the end-effector frame $\mathbf{T}_{WE}^t \in \mathrm{SE}(3)$ and $g^t \in \{0, 1\}$ indicates the gripper state, where $0$ and $1$ represent open and closed status, respectively.
Specific to simulation, we add a list of object identified by id $\xi_{sim} = \{x_1, x_2, \dots, x_n\}$ to $d$ for masking. Moreover, in order to dependency on ground-truth trackers, the real-world demonstrations omit these identifiers, relying instead on the VLM-based object grounding described in our task planning module.

\subsection{Module 1: Scene-Relevant Task Generation}
\label{subsec:task_generation}

\begin{figure}[t!]
    \centering
    \includegraphics[width=0.89\linewidth]{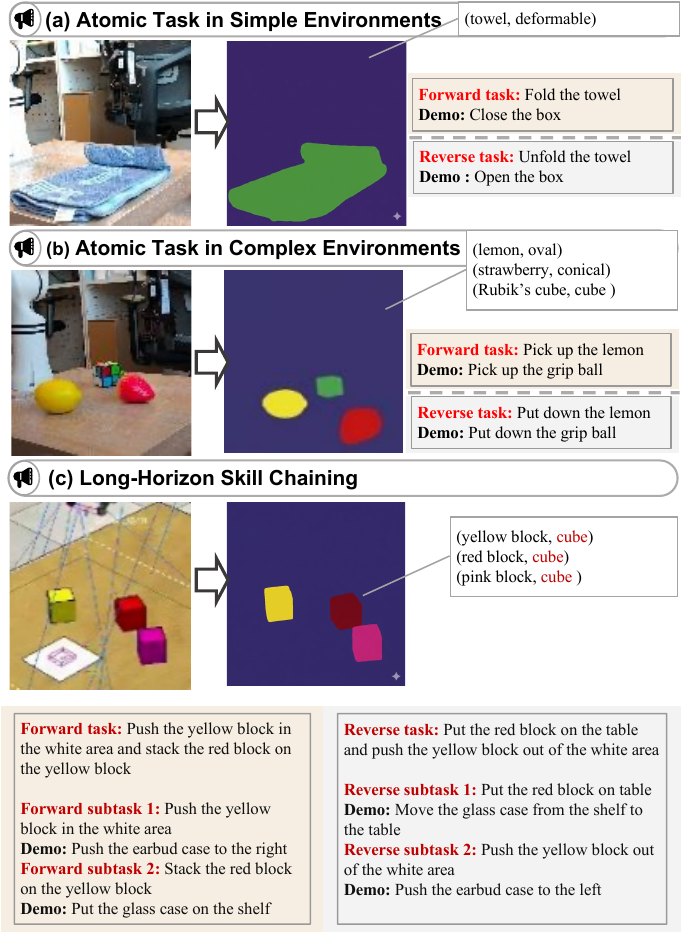}
    \caption{Overview of our hierarchical Scene-Relevant Task Generation framework across varying complexities. \textbf{(a) Atomic Task in Simple Environments:} The VLM performs direct affordance matching, mapping a deformable object task (folding a towel) to a geometrically congruent 3D prior (closing a box). \textbf{(b) Atomic Task in Complex Environments:} Through selective attention, the planner actively masks out distractors (\textit{e.g.}, strawberry, Rubik's cube) to precisely ground the target object (lemon) and retrieve a robust prior. \textbf{(c) Long-Horizon Skill Chaining:} For multi-step tasks, the VLM orchestrates a forward skill chain (pushing and stacking blocks) while concurrently generating a strict Last-In, First-Out (LIFO) causal sequence to autonomously construct executable environment-resetting plans.}
    \label{fig:task_plan_demo}
\end{figure}

As illustrated in Fig. \ref{fig:task_plan_demo}, to generate physically feasible and semantically meaningful tasks in unstructured environments, we propose a hierarchical visual task planning framework. To leverage the reasoning capabilities of VLM (denoted as $\mathcal{M}$), we propose a 2-stage process: (1) \textbf{semantic object grounding}, where we utilize VLM to ground relevant objects; (2) \textbf{hierarchical task planning}, where we translate visual observations into structured task plans.

\subsubsection{Semantic Object Grounding}
To get an accurate understanding of the scene, instead of feeding raw images directly into the planning module (which may lead to hallucinations), we first employ a VLM-based object detector to extract a structured semantic representation of the current scene. The process can be formulated as follows:
\begin{equation}
\xi_{scene} = \mathcal{M}(p_{ground}, s_0)
\end{equation}
Given an RGB image of the initial scene $s_0$ with a grounding language prompt $p_{ground}$, the pipeline queries the VLM to identify all candidate objects and their geometric attributes (\textit{e.g.}, identifying a lemon as ``oval'' or a strawberry as ``conical'', as shown in Fig. \ref{fig:task_plan_demo}b). We denote this structured output as $\xi_{scene} = \{x_1, x_2, \dots, x_n\}, x_i = (name, shape)$. This output serves as a hard constraint for subsequent planning, ensuring the robot only attempts to interact with objects physically present in the workspace.

\subsubsection{Hierarchical Task Planning}
We formalize the hierarchical task planning as a conditional generation problem. Given the current scene observation $s$ and a task-level prompt $p_{level}$, the VLM $\mathcal{M}$ generates a structured response $y$:

\begin{equation}
y = \mathcal{M}(p_{level}, s)
\end{equation}

where $y$ encapsulates a set of task-relevant objects $\xi_{mask}$ for precise visual grounding and a sequence of executable subtasks $\mathcal{T} = \{(a_t, d_t, c_t)\}_{t=1}^T$. For each step $t$, the VLM planner identifies an atomic action $a_t$, retrieves the most semantically and geometrically congruent demonstration $d_t$ from the affordance library $L$, and provides a textual description $c_t$ to guide the downstream imitation learning policy.

Based on scene complexity and task horizon, our planner dynamically adapts across three scenarios, as comprehensively detailed in Fig. \ref{fig:task_plan_demo}:
1) \textit{Atomic Task in Simple Environments}: For uncluttered scenes, the VLM performs \textbf{Direct Affordance Matching} (Fig. \ref{fig:task_plan_demo}a). For example, it maps the task of ``folding a deformable towel'' directly to the geometrically similar demonstration of ``closing a box'', thereby generating its reverse task (``unfolding the towel'') based on the ``opening a box'' demonstration. 
2) \textit{Atomic Task in Complex Environments}: When distractors are present, a specific prompt enforces \textbf{Selective Attention} (Fig. \ref{fig:task_plan_demo}b). The VLM identifies a target subset $\xi_{mask}$ (\textit{e.g.}, focusing purely on the target lemon while actively ignoring the strawberry and Rubik's cube) to prevent hallucinated interactions, retrieving the ``pick up grip ball'' skill as a robust 3D prior. 
3) \textit{Long-Horizon Skill Chaining}: For multi-step tasks, the VLM logically chains atomic skills (Fig. \ref{fig:task_plan_demo}c) without decomposing them into smaller primitives. The system prompt enforces physical causality and the simultaneous generation of environment-resetting plans. For example, generating a forward sequence (``push the yellow block into the white area'', ``stack the red block'') concurrently produces a strict Last-In, First-Out (LIFO) constrained reverse sequence (``put the red block on the table'', ``push the yellow block out''). This explicitly ensures the multi-step plan is both physically executable and inherently self-reversible.

\subsubsection{In-Context Skill Retrieval via Affordance Library}
To bridge high-level planning with low-level control, both atomic and long-horizon planning modes utilize an in-context retrieval mechanism. For every generated subtask, the VLM estimates a similarity rank based on action semantics and object geometry, returning an ordered list of reference demonstrations $\mathcal{D}=\{d_{1},d_{2},...,d_{r}\}$. Specifically, our prompting framework instructs the VLM to evaluate similarities across two distinct dimensions: \textbf{Action Similarity} (\textit{e.g.}, ensuring the motion trajectories of folding and closing align) and \textbf{Geometric/Functional Similarity} (\textit{e.g.}, recognizing that a ``lemon'' is geometrically congruent to an oval ``grip ball''). This dual-criteria retrieval eliminates the need for hallucinating arbitrary 3D waypoints, allowing the downstream imitation learning policy to leverage robust geometric priors for zero-shot generalization to novel objects.

\subsection{Module 2: Task Execution via In-Context Imitation Learning}
\label{subsec:task_execution}

Inspired by Instant Policy~\cite{vosylius2024instantpolicy}, RADAR employs
graph-based In-Context Imitation Learning (ICIL) to execute novel subtasks
without task-specific fine-tuning. Given the demonstrations retrieved by the
planning module, the policy jointly represents the demonstration context
$\mathcal{G}_{c}$, the current segmented point-cloud observation
$\mathcal{G}_{l}^{t}$, and the predicted future actions
$\mathcal{G}_{l}^{a}(a)$ as a heterogeneous graph:
\begin{equation}
\mathcal{G}^{k}
=
\mathcal{G}\left(
\mathcal{G}_{l}^{a}(a^{k}),
\mathcal{G}_{c},
\mathcal{G}_{l}^{t}
\right)
\end{equation}
The demonstration graphs encode object geometry, end-effector poses, gripper
states, and temporal motion, providing the policy with a 3D behavioral prior
for the current task.

Action generation is formulated as a conditional reverse-diffusion process.
Starting from Gaussian noise $a^{K}\sim\mathcal{N}(0,I)$, a heterogeneous
graph transformer predicts the denoising direction for the future action
nodes while keeping the demonstration and observation context fixed. One
reverse step is written as
\begin{equation}
\begin{aligned}
\mathcal{G}^{k-1}
=
\mathcal{G}\Big(
\mathcal{G}_{l}^{a}\big(
\alpha_k(
a^{k}
-
\gamma_k\epsilon_{\theta}(\mathcal{G}^{k},k))
+
\sigma_k\mathbf{z}
\big),
\mathcal{G}_{c},
\mathcal{G}_{l}^{t}
\Big)
\end{aligned}
\label{eq:instant_policy_denoising}
\end{equation}
where $\mathbf{z}\sim\mathcal{N}(0,I)$ and
$\alpha_k$, $\gamma_k$, and $\sigma_k$ are determined by the diffusion
schedule. The predicted gripper-keypoint updates are converted into valid
$\mathrm{SE}(3)$ end-effector transformations through rigid alignment. After
$K$ denoising steps, the resulting action $a^{0}$ is executed, and the policy
is repeatedly queried with updated observations to provide closed-loop
control.

\subsection{Module 3: Automated Success Evaluation}
\label{subsec:success_eval}
To close the loop of our autonomous data collection pipeline, it is imperative to evaluate the outcome of the executed policy without requiring human-in-the-loop verification. To this end, we introduce an \textbf{Automated Success Evaluation} module. This module leverages a Vision-Language Model (VLM) to visually inspect the post-execution workspace and determine whether the physical state aligns with the semantic goal of the commanded subtask. 

To mitigate the inherent variability and conversational verbosity of raw VLM outputs, we formulate the success detection as a structured, three-stage Visual Question Answering (VQA) pipeline: (1) Semantic Task-to-Query Translation, (2) Vision-Language Assessment, and (3) Robust Boolean Decoding.

\subsubsection{Semantic Task-to-Query Translation}
Standard VLMs often struggle to evaluate imperative task commands (\textit{e.g.}, ``put the yellow ball on the blue plate'') directly. To optimize the VLM's reasoning, we first translate the task description $c_t$ into a targeted, interrogative VQA query $q_{vqa}$. 

We employ a Large Language Model (LLM), denoted as $\mathcal{M}_{trans}$, guided by a few-shot in-context prompt containing diverse manipulation examples (\textit{e.g.}, mapping ``move the red object from the cloth to the table'' to ``Is the red object on the cloth or the table?''). This process translates the action-oriented command into a state-oriented visual query:
\begin{equation}
q_{vqa} = \mathcal{M}_{trans}(c_t, p_{vqa})
\end{equation}
where $p_{vqa}$ represents the few-shot prompt template designed to elicit specific spatial and relational questions.

\subsubsection{Vision-Language Assessment}
Following the execution of the policy over the temporal horizon $T$, the robot captures the final visual observation of the workspace, denoted as $s_T$. The generated query $q_{vqa}$ and the image $s_T$ are subsequently fed into a state-of-the-art VLM (\textit{e.g.}, GPT-4V or CogVLM), denoted as $\mathcal{M}_{vlm}$. The VLM acts as an embodied evaluator, analyzing the spatial relationships and object states within $s_T$ to generate a raw textual assessment $r_{vlm}$:
\begin{equation}
r_{vlm} = \mathcal{M}_{vlm}(s_T, q_{vqa})
\end{equation}

\subsubsection{Robust Boolean Decoding}
Since the raw response $r_{vlm}$ from the VLM may contain auxiliary reasoning or conversational fillers (\textit{e.g.}, ``Yes, I can see that the object is...'' instead of a strict boolean), directly parsing this output is error-prone. To ensure a deterministic control flow for our autonomous pipeline, we introduce a final decoding step. 

We utilize a parsing LLM $\mathcal{M}_{parse}$ to distill the verbose evaluation into a strict binary success signal $b_{succ} \in \{True, False\}$. The parser is conditioned on the original task $c_t$, the generated query $q_{vqa}$, and the VLM's response $r_{vlm}$:
\begin{equation}
b_{succ} = \mathcal{M}_{parse}(c_t, q_{vqa}, r_{vlm})
\end{equation}
Instead of relying on heuristic error handling, this extracted binary signal $b_{succ}$ directly serves as the deterministic trigger for state transitions in our downstream pipeline. These signals dictate whether the system advances to the reverse resetting module or aborts to initiate a new planning cycle, as detailed in Section \ref{subsec:env_reset}. Furthermore, these discrete success metrics are logged to maintain historical task statistics, enabling the system to track the reliability of specific skills over time.

\subsection{Module 4: Autonomous Environment Reset}

\label{subsec:env_reset}
To achieve truly continuous and autonomous data collection, minimizing human intervention between episodes is critical. We introduce an \textbf{Autonomous Environment Reset} mechanism, which leverages the reasoning capabilities of the VLM to automatically restore the workspace to its initial state $s_0$ following the completion of a generated task. Instead of relying on hard-coded or heuristic reset scripts, our framework formulates the environment reset as an inverse task planning problem tightly coupled with the forward generation.

\subsubsection{Simultaneous Forward-Reverse Planning}
During the hierarchical task planning phase, the VLM is prompted to act as a causal reasoning engine. It simultaneously generates the primary executable plan (denoted as the \textit{forward task}) and its exact causal inverse (denoted as the \textit{reverse task}). 

For atomic tasks, the VLM directly infers the inverse affordance based on the object's geometric properties. For example, if the forward atomic action is ``pick up the cup'', the VLM proposes ``place the cup down'' as the reverse task and retrieves the most semantically congruent demonstration from the affordance library $L$.

For long-horizon multi-step tasks, the environment reset strictly adheres to a Last-In, First-Out (LIFO) causal sequence constraint. Given a forward plan consisting of $N$ subtasks $\mathcal{T}_{fwd} = \{(a_i^f, d_i^f, c_i^f)\}_{i=1}^N$, the VLM constructs a reverse plan $\mathcal{T}_{rev} = \{(a_j^r, d_j^r, c_j^r)\}_{j=1}^N$. Crucially, each reverse subtask at step $j$ must explicitly undo the physical state changes introduced by the forward subtask at step $i = N-j+1$. This ensures physical feasibility during the restoration phase (\textit{e.g.}, a box must be opened before the cube placed inside it can be extracted).

\subsubsection{Continuous Collection via Finite State Machine}
To seamlessly integrate physical actions and data logging into a continuous, human-out-of-the-loop pipeline, we orchestrate the process using a Finite State Machine (FSM), as visualized in Fig. \ref{fig:whole_logic}. To ensure logical clarity, our FSM explicitly decouples the physical execution states—Task Planning (\textbf{State A}), Forward Execution (\textbf{State B}), and Reverse Execution (\textbf{State C})—from the concurrent data routing actions—Dual Storage (\textbf{State D}) and Single Storage (\textbf{State E}).

Governed by the binary success signal $b_{succ}$ from the VQA evaluator, the system executes the following operational loops, with data storage acting as triggered side-effects:

\begin{itemize}
    \item \textbf{Continuous Success Loop (B $\to$ C $\to$ B):} If both the forward task and the reverse reset succeed, the physical environment is perfectly restored. The control flow seamlessly loops from State C directly back to State B to \textbf{repeatedly execute the same assigned task}. Concurrently, this successful cycle triggers the \textbf{Dual Storage (State D)} mechanism, validating and saving both the forward and reverse trajectories. This cyclic execution allows the system to continuously harvest diverse trajectory variations of a specific skill without the computational overhead of VLM re-planning.
    
    \item \textbf{Asymmetric Recovery Loop (B $\to$ C $\to$ A):} If the forward task succeeds but the reverse reset fails, the physical environment is left un-restored. To recover, the control flow forcibly breaks the loop and routes back to Task Planning (State A), treating the altered workspace as a novel initial scene. Concurrently, this transition triggers the \textbf{Single Storage (State E)} mechanism, selectively retaining the perfectly valid forward trajectory while discarding the failed reset attempt.
    
    \item \textbf{Forward Abort (B $\to$ A):} If the initial forward execution fails, the invalid trajectory is immediately discarded (no storage triggered). The control flow aborts the current execution queue and transitions back to State A to re-observe and re-plan.
\end{itemize}

As depicted in the global architecture (Fig. \ref{fig:whole_logic}), this state-decoupled orchestration guarantees the continuous streaming of high-fidelity data into the Affordance Library. By strictly separating the execution loops from the asymmetric data routing, the pipeline effectively bypasses reset failures and eliminates the need for manual scene restoration, transforming it into a truly self-sustaining engine.

\section{EXPERIMENTS}
\label{sec:experiments}

\subsection{Experimental Setup}
\label{subsec:exp_setup}
We evaluate our automated pipeline in the RLBench simulation \cite{james2020rlbench}. We select 7 atomic tasks and 3 complex long-horizon tasks, comparing our system against two state-of-the-art vision-language manipulation baselines: MOKA \cite{fang2024moka} and ReKep \cite{huang2024rekep}. 

\textbf{Implementation Details \& Evaluation Protocol:} To rigorously isolate and evaluate the forward task generation and execution capabilities, the environments are reset using simulation ground truth between rollouts. We report the success rate over 10 independent rollouts per task with randomized initial states. Empirically, we provide a single high-quality demonstration (1-shot) as the context, as increasing demonstrations did not yield proportional gains. Furthermore, we utilized a VLM instead of CLIP for skill retrieval, as our preliminary tests showed CLIP embeddings are heavily noun-biased and fail to distinguish fine-grained actionable semantics.


\begin{table}[htbp]
\centering
\caption{Success rates in RLBench over 10 rollouts per task.}
\label{tab:sim_results}
\footnotesize
\setlength{\tabcolsep}{3.2pt}
\renewcommand{\arraystretch}{1.10}
\begin{tabular}{@{}clccc@{}}
\toprule
\textbf{Type} & \textbf{Task} & \textbf{ReKep} & \textbf{MOKA} & \textbf{Ours} \\
\midrule

\multirow{7}{*}{\textit{Atomic}}
& Large Container (Cup)       & 2/10 & 2/10 & \textbf{8/10}  \\
& Large Container (Block)     & 0/10 & 3/10 & \textbf{8/10}  \\
& Large Container (Laptop)    & 0/10 & 1/10 & \textbf{9/10}  \\
& Push Block                  & 4/10 & 4/10 & \textbf{10/10} \\
& Stack Block                 & 4/10 & 1/10 & \textbf{8/10}  \\
& Close Box                   & 4/10 & 3/10 & \textbf{10/10} \\
& Open Box                    & 2/10 & 2/10 & \textbf{7/10}  \\

\addlinespace[2pt]
\cmidrule(lr){1-5}

\multirow{3}{*}{\shortstack[c]{\textit{Long-}\\\textit{horizon}}}
& Put Laptop \& Cup into Tray & 1/10 & 0/10 & \textbf{8/10} \\
& Push \& Stack Blocks        & 0/10 & 0/10 & \textbf{4/10} \\
& Close then Open Box         & 2/10 & 1/10 & \textbf{9/10} \\
\bottomrule
\end{tabular}
\end{table}

\subsection{Main Results}
\label{subsec:main_results}
\begin{figure}[t]
    \centering
    \includegraphics[width=0.8\linewidth]{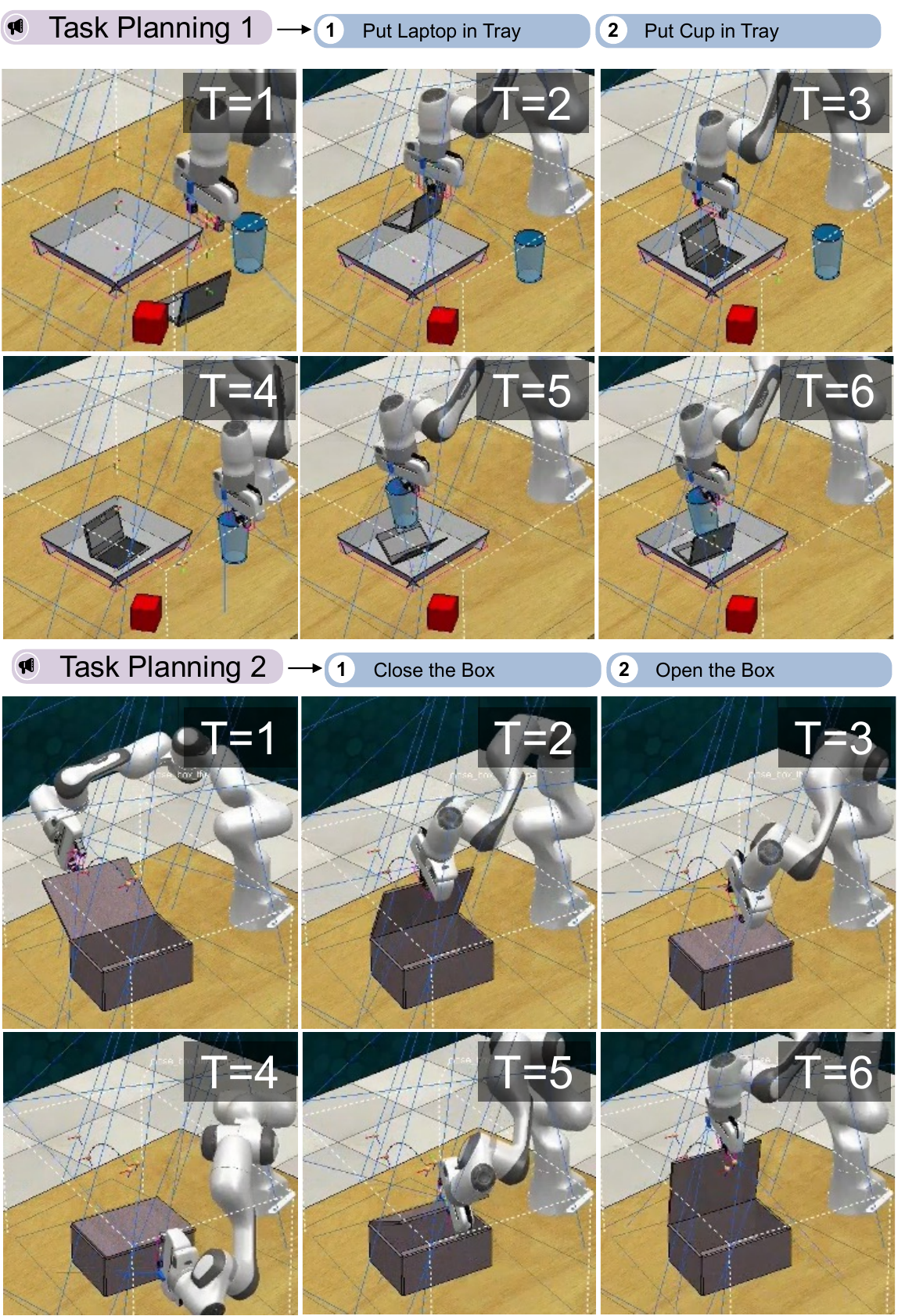}
    \caption{Visualizations of long-horizon tasks in the RLBench simulation. \textbf{(a) Put Laptop \& Cup into Tray:} The robot successfully executes a multi-step sequence requiring interactions with multiple distinct objects. \textbf{(b) Close then Open Box:} The pipeline reliably performs state-dependent articulated actions, demonstrating its robust skill chaining capability.}
    \label{fig:sim_long_horizon}
\end{figure}

As shown in Table \ref{tab:sim_results}, our pipeline significantly outperforms ReKep and MOKA. While baselines exhibit moderate performance on simple atomic tasks, their success rates plummet to near-zero in long-horizon scenarios (\textit{e.g.}, ``Push \& Stack Blocks'', as illustrated in Fig. \ref{fig:experiments}c). In contrast, our system maintains robust performance, achieving 80\%-90\% success rates on demanding multi-step tasks. Specifically, the pipeline effectively orchestrates sequential skill chains and dependent articulated actions (\textit{e.g.}, opening a previously closed box, as visualized in Fig. \ref{fig:sim_long_horizon}), proving its capability to autonomously gather high-quality data for complex behaviors.

\subsection{Ablation: The Necessity of Point Cloud Masking}
\label{subsec:ablation}

\begin{table}[htbp]
\centering
\caption{Ablation on Point Cloud Masking}
\label{tab:masking_ablation}
\begin{tabular}{lcc}
\toprule
\textbf{Task} & \textbf{W/o Masking} & \textbf{Ours (With Masking)} \\
\midrule
Large Container (Cup)   & 0.10 & \textbf{0.80} \\
Large Container (Block) & 0.00 & \textbf{0.80} \\
Push Block              & 0.00 & \textbf{1.00} \\
\bottomrule
\end{tabular}
\end{table}

To evaluate our VLM-driven selective attention, we ablate the masking module and feed raw scene point clouds into the execution policy. As shown in Table \ref{tab:masking_ablation}, removing semantic masking causes catastrophic failures, with success rates for grasping and pushing dropping to near zero due to susceptibility to distractors. This highlights that explicit semantic masking is crucial for execution robustness in cluttered scenes (a real-world qualitative example is depicted in Fig. \ref{fig:experiments}b).

\subsection{Real-World Deployment and Limitations}
\label{subsec:real_world}

\begin{figure}[t]
    \centering
    \includegraphics[width=0.8\linewidth]{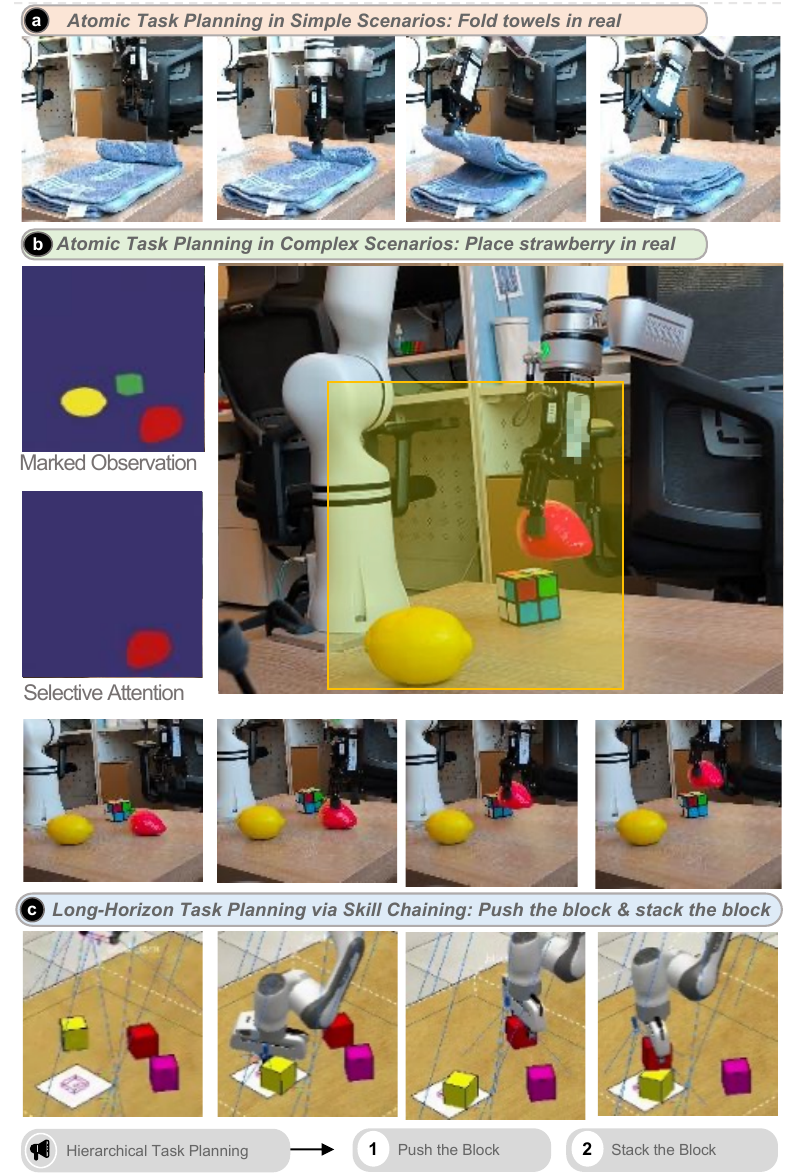}
    \caption{Qualitative results of our automated data collection pipeline across different scenarios. \textbf{(a) Simple Atomic Scenario:} The robot executes deformable object manipulation (folding a towel) in a real-world setting without distractors. \textbf{(b) Complex Atomic Scenario:} The pipeline utilizes selective attention to isolate a target object (a strawberry) among visual distractors for precise grasping. \textbf{(c) Long-Horizon Scenario:} In simulation, the VLM decomposes a complex instruction into a sequential skill chain (\textit{e.g.}, first pushing, then stacking a block).}
    \label{fig:experiments}
\end{figure}

To validate practical applicability, we deployed RADAR on a physical Realman RM65-B arm with a RealSense D435i camera, using SAM \cite{kirillov2023segment} and XMem++ \cite{bekuzarov2023xmem} for real-time 3D object segmentation. 

\textbf{Qualitative Feasibility:} Operating under a 1-shot adaptation paradigm without domain-specific fine-tuning, the robot successfully executed intricate tasks. In simple environments, it handled deformable object manipulation (\textit{e.g.}, folding a towel, Fig. \ref{fig:experiments}a). In complex scenarios, it utilized selective attention for precise grasping among distractors (\textit{e.g.}, targeting a strawberry, Fig. \ref{fig:experiments}b). These qualitative results strongly indicate RADAR's immense potential as a scalable, human-out-of-the-loop engine for physical robot learning.

\textbf{Limitations and Future Work (The Reset Challenge):} While our pipeline demonstrates the feasibility of autonomous data generation, fully 100\% reliable environment resetting remains an open challenge. Probabilistically, chaining forward execution with a causal reverse reset inherently compounds failure rates ($p_{total} \approx p_{forward} \times p_{reverse}$). Currently, our FSM acts as a robust proof-of-concept for simple-to-moderate scenes. Overcoming this compounding error in highly unstructured environments—perhaps via multi-modal tactile feedback or high-frequency visual servoing—represents an exciting frontier for future work.

\section{Conclusion}
\label{sec:conclusion}

We present \textbf{R}obust \textbf{A}utonomous \textbf{D}ata \textbf{A}cquisition for \textbf{R}obotics (\textbf{RADAR}), a self-sustaining, human-out-of-the-loop data generation engine. Orchestrated by a decoupled Finite State Machine (FSM), RADAR achieves a seamless brain-cerebellum synergy by coupling VLM-based cognitive planning and LIFO-constrained autonomous environment resetting with the sub-millimeter precision of GNN-based in-context imitation learning. Anchored by 3D human demonstrations, it circumvents the geometric hallucinations of purely 2D pipelines, while its asymmetric data routing guarantees continuous data streaming despite execution failures.

RADAR achieves up to 90\% success on complex simulated tasks and reliably executes contact-rich physical skills via few-shot adaptation without domain-specific fine-tuning. Looking ahead, our future work will proceed along two exciting frontiers: tackling the open challenge of compounding reset errors via multi-modal sensory integration, and leveraging the generated high-fidelity datasets to train downstream foundational visuomotor policies (\textit{e.g.}, Diffusion Policy) for dynamic real-world environments.











\end{document}